%% file: main1.tex
\documentclass{article}

\usepackage[preprint]{arxiv_preprint}

\usepackage{times}
\usepackage{microtype}
\usepackage{hyperref}
\usepackage{url}
\usepackage{amsmath,amssymb,amsfonts}
\usepackage{bm}
\usepackage{booktabs}
\usepackage{graphicx}
\usepackage{xcolor}
\usepackage{enumitem}
\usepackage[capitalize,noabbrev]{cleveref}
\usepackage{multirow}
\input{math_commands}

\usepackage{colortbl}
\definecolor{bestcolor}{RGB}{250, 127, 111}
\newcommand{\best}[1]{\textbf{#1}}

\newcommand{\second}[1]{\underline{#1}}
\title{A World Model of Radiologist Reading for Medical Image Representation Learning}

\author{
  Yiwei Li\thanks{Equal contribution.} \\
  University of Georgia \\
  \And
  Zihao Wu\footnotemark[1] \\
  University of Georgia \\
  \And
  Huaqin Zhao \\
  University of Georgia \\
  \And
  Yifan Zhou \\
  University of Georgia \\
  \And
  Chao Cao \\
  University of Texas at Arlington \\
  \And
  Dajiang Zhu \\
  University of Texas at Arlington \\
  \And
  Tianming Liu \\
  University of Georgia \\
  \And
  Lin Zhao\thanks{Corresponding author.} \\
  New Jersey Institute of Technology \\
}

\begin{document}
\maketitle
\begin{abstract}
Radiologist eye-tracking data provide a rich record of how experts search, compare, and accumulate evidence during image reading; yet, existing methods exploit this signal only partially, either as a static spatial prior or as an auxiliary prediction target decoupled from diagnosis. We propose GazeWorld, a medical imaging world model that treats the image as the world and the radiologist's fixation sequence as a trajectory through it. GazeWorld autoregressively predicts the latent representation of the next fixated patch from all previously visited ones, while a spatial-completion branch covers unvisited regions. At inference, GazeWorld generates a sequence of patch representations from the image alone without requiring real gaze data. Frozen GazeWorld features achieve state-of-the-art diagnostic accuracy across all nine supervised settings on CheXpert, RSNA Pneumonia, and SIIM-ACR Pneumothorax, as well as the highest zero-shot accuracy on all three benchmarks. On the GazeSearch benchmark, a generic decoder trained on the same frozen features outperforms the purpose-built LogitGaze-Med by over 16\% in ScanMatch and 22\% in SED, despite not being explicitly trained to predict gaze. GazeWorld demonstrates that modeling how experts read, not just what they conclude, offers a promising pretraining paradigm for medical imaging AI.
\end{abstract}

\input{introduction}
% §1
\input{related_work}     % §2
\input{method}           % §3
\input{experiments}     % §4
\input{conclusion}     % §5  (to be written)

\bibliographystyle{unsrtnat}
\bibliography{references}
\input{appendix}

\end{document}

%% file: math_commands.tex
\usepackage{amsmath,amssymb,amsfonts}
\usepackage{bm}

%% file: introduction.tex
% ============================================================
%  §1  Introduction
% ============================================================

\section{Introduction}
\label{sec:intro}

Two dark clouds loom over the clinical translation of medical imaging AI. Large-scale medical imaging datasets remain scarce because acquisition is bound by clinical access, cross-institutional sharing is curtailed by privacy regulations, and expert annotations are prohibitively expensive to obtain at scale~\citep{irvin2019chexpert,johnson2019mimic,rajpurkar2017chexnet}. Without sufficiently large and diverse training sets, models readily overfit to dataset-specific shortcuts and degrade under distribution shifts~\citep{azizi2021big,sowrirajan2021moco}. Compounding this data bottleneck, most current models emit a diagnostic label without exposing the evidence behind the
prediction~\citep{chen2024chexagent,zhou2023foundation}, leaving
clinicians unable to assess whether the output reflects sound
reasoning or a spurious shortcut and ultimately eroding trust required for clinical deployment.

Radiologist gaze offers a promising inroad into both challenges. Eye-tracking data encode expert prior knowledge about where diagnostically relevant evidence lies, providing a form of supervision that can reduce reliance on large labeled datasets; several studies have incorporated gaze as a spatial attention prior and demonstrated improved performance~\citep{shentu2024cxr,wang2022follow,ma2024eye,wang2025improving}, though collapsing eye movements into static heatmaps discards the order in which evidence was gathered. Beyond this, gaze trajectories record the temporal sequence in which a radiologist searches, compares, and accumulates evidence, offering an observable trace of diagnostic reasoning that can guide models toward interpretable decisions. Recent work has shown that such trajectories can be predicted from chest radiographs alone~\citep{pham2025gazesearch,lvovhuman}, but the predicted scanpaths remain decoupled from the diagnosis. Both lines leave the temporal structure of radiologist evidence gathering unused for diagnosis.

We propose to close this gap by formulating a world model of the radiologist's diagnostic process. When reading a medical image, a radiologist navigates a perceptual world defined by the image. At each fixation, the radiologist extracts local evidence and selects the next region of interest conditioned on prior observations, iterating until sufficient evidence supports a diagnosis. We formalize this as a world model that predicts the latent representation of each successive fixation, conditioned on the current fixation and accumulated evidence. Each fixation along the trajectory thus becomes a prediction target, yielding dense self-supervised signals from each image without requiring diagnostic labels. At the same time, the learned representation retains the sequential structure of expert reasoning, offering a basis for interpretability that static attention maps cannot provide.

We instantiate this formulation as GazeWorld, a medical imaging world model that captures the sequential structure of expert reading through gaze-ordered representation prediction. A visual encoder maps the image into patch-level representations and an autoregressive predictor, following the joint-embedding predictive architecture (JEPA)~\citep{lecun2022path,assran2023self}, processes the gaze-ordered fixation sequence to predict the latent representation of the next fixated patch from all previously visited ones. In parallel, a spatial-completion branch uses the full gaze context to predict the representations of all unvisited patches. At inference, the autoregressive predictor generates the representations from the image alone without requiring gaze data for downstream tasks.

We evaluate GazeWorld on chest radiograph diagnosis and scanpath prediction. For diagnosis, linear probes on frozen GazeWorld features achieve state-of-the-art accuracy across all nine supervised settings on CheXpert~\citep{irvin2019chexpert}, RSNA Pneumonia~\citep{colak2021rsna}, and SIIM-ACR Pneumothorax~\citep{siim_acr_pneumothorax_2019}, as well as the highest zero-shot accuracy on all three benchmarks. For scanpath prediction, a generic decoder trained on the same frozen features outperforms the purpose-built LogitGaze-Med~\citep{lvovhuman} by over 16\% in ScanMatch and 22\% in SED on the GazeSearch benchmark, despite not being explicitly trained to predict gaze.

Our contributions are as follows:

\begin{itemize}
\item We introduce the concept of a gaze world model for medical imaging, which formulates the radiologist's reading process as a trajectory through a perceptual world defined by the image. This formulation fully exploits the temporal structure of expert gaze as a self-supervised training signal, moving beyond static heatmaps and decoupled scanpath prediction.
\item We instantiate this formulation as GazeWorld, which combines gaze-ordered autoregressive prediction in latent space with a spatial-completion branch for unvisited regions, shaping the representation to capture both diagnostic content and sequential reading structure.
\item GazeWorld achieves state-of-the-art diagnostic accuracy across all nine supervised settings and the highest zero-shot accuracy on three chest radiograph benchmarks. A generic decoder on the same frozen features outperforms purpose-built scanpath predictors, showing that a single representation supports both accurate diagnosis and interpretable scanpath decoding.

\end{itemize}

%% file: related_work.tex
\section{Related Work}
\label{sec:related}

\subsection{Gaze Supervision and Scanpath Prediction}

Radiologist gaze has been used in medical AI mainly in two ways. The first
uses gaze as spatial supervision for diagnosis or representation learning,
typically by converting fixations into heatmaps, attention maps, or relation
matrices~\citep{wang2022follow,saab2021observational,stember2019eye,
hsieh2024eyexnet}. These signals provide weak but clinically meaningful
localization cues, helping models attend to regions that radiologists consider
diagnostically relevant. For example, EGMA~\citep{ma2024eye} uses
fixation--derived attention to guide fine--grained image--text alignment, while
eCLIP~\citep{kumar2024improving} and other gaze--guided methods use expert
attention to regularize contrastive learning, region--aware prediction, or
diagnostic reasoning. These approaches demonstrate that gaze can improve
clinical relevance and reduce reliance on dense manual annotations. However,
they mostly treat gaze as static spatial supervision, preserving where
radiologists looked while discarding the temporal order of evidence gathering,
comparison, and integration.

The second line models gaze as a scanpath prediction problem.
GazeFormer~\citep{mondal2023gazeformer}, HAT~\citep{yang2024unifying},
MedGaze~\citep{awasthi2025modeling}, GazeSearch~\citep{pham2025gazesearch},
and LogitGaze--Med~\citep{lvovhuman} predict fixation trajectories or dwell
times from medical images. These methods explicitly preserve gaze order and
evaluate human--like visual search, but scanpath prediction is usually treated
as a standalone output task. In contrast, GazeWorld uses radiologist scanpaths
to define the autoregressive prediction order for representation learning, so
the temporal structure of expert search directly shapes the visual encoder.

\subsection{World Models and Representation--Space Prediction}

World models aim to learn predictive structure in latent space as observations
accumulate~\citep{lecun2022path}. Instead of reconstructing pixels, JEPA--style
methods predict abstract representations of unseen regions from visible
context, which makes the objective less sensitive to low--level appearance noise
and more focused on semantic structure~\citep{assran2023self,bardes2023v}.
Autoregressive vision models provide another predictive formulation by
modeling visual tokens sequentially, but their prediction order is usually
fixed by design, such as raster scan or predefined token layouts
~\citep{chen2020generative,el2024scalable}. These predictive objectives are
especially appealing for medical imaging, where diagnostic signals are subtle,
spatially distributed, and often better captured in representation space than
at the pixel level. Recent medical models such as
CheXWorld~\citep{yue2025chexworld} and
US--JEPA~\citep{radhachandran2026us} show that representation--space prediction
can improve medical visual features by modeling anatomy, global layout, and
domain variation.

However, existing predictive structures are typically defined by random masks,
block masks, crops, or raster order. These choices are convenient but not tied
to clinical reading behavior. GazeWorld instead uses radiologist fixation
sequences as the prediction structure, where previously fixated patches form
the context and the next expert--fixated patch is the autoregressive target.
This grounds representation--space prediction in the temporal process of
diagnostic evidence accumulation.

\subsection{Medical Vision Representation Learning}

Medical vision models rely on supervised pretraining, self--supervised
learning, contrastive learning, and vision--language pretraining to reduce
annotation cost and improve transfer~\citep{irvin2019chexpert,
johnson2019mimic,rajpurkar2017chexnet,sowrirajan2021moco,azizi2021big,
xiao2023delving}. Representative medical vision--language models include
GLoRIA~\citep{huang2021gloria}, MedCLIP~\citep{wang2022medclip},
BioViL~\citep{bannur2023learning}, MGCA~\citep{wang2022multi},
MedKLIP~\citep{wu2023medklip}, and PRIOR~\citep{cheng2023prior}, which
learn transferable representations from image labels, unlabeled images, or
image--text pairs.

GazeWorld is complementary to these pretraining paradigms rather than a
replacement for them. It can be applied on top of existing medical visual
backbones, such as MGCA, EGMA, or MedCLIP, to further refine their
representations with expert gaze dynamics. Thus, our contribution is not a new
foundation pretraining pipeline from scratch, but a gaze-ordered world-model
refinement objective that injects radiologist reading behavior into already
pretrained medical encoders.

%% file: method.tex
% ============================================================
%  §3 Method — GazeWorld with Spatial Completion
% ============================================================
\begin{figure*}[h]
  \centering
  \includegraphics[width=\textwidth]{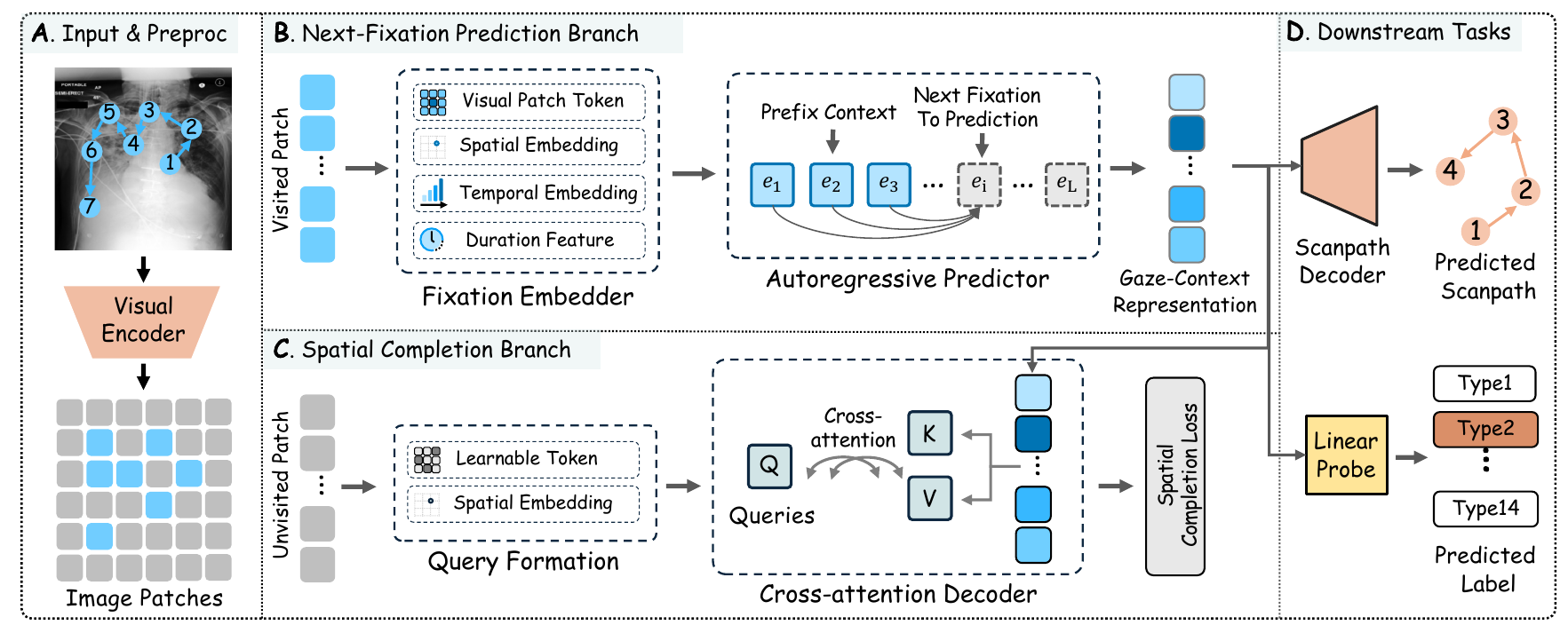}
 \caption{
   \textbf{GazeWorld overview.}
After the chest radiograph and radiologist fixation sequence are processed in
Part A, the image is converted into patch-level semantic tokens and the patch
grid is divided into visited and unvisited regions. In Part B, visited patch
tokens are combined with spatial, temporal, and duration information by the
fixation embedder, and an autoregressive predictor performs latent-space prediction of the next expert-fixated representation from the prefix context. The resulting gaze-context
representation is then used in Part C, where unvisited patch queries are formed
from learnable tokens and spatial embeddings and completed through a
cross-attention decoder. Finally, Part D evaluates the representations via scanpath decoding and diagnostic linear probing.
}
  \label{fig:architecture}
\end{figure*}

\section{Method}
\label{sec:method}

GazeWorld casts the medical image as a perceptual world and the radiologist's fixation sequence as a trajectory through it. An autoregressive predictor processes the gaze-ordered fixation sequence and predicts the latent representation of the next fixated patch from all previously visited ones, turning each fixation into a self-supervised training signal. To exploit unvisited regions, a spatial-completion branch uses the full gaze context to predict all unvisited patch representations in parallel. The final pretraining objective combines next-fixation prediction with gaze-conditioned spatial completion.

% ------------------------------------------------------------
\subsection{Patch Representations and Fixation Sequence}
\label{sec:method:tokens}
% ------------------------------------------------------------

Following the input and preprocessing stage in Fig.~\ref{fig:architecture}A,
let $\mathbf{X}$ denote a medical image. An online visual encoder
$\phi$ maps $\mathbf{X}$ into a set of patch-level semantic
representations:
\begin{equation}
  \mathbf{Z}
  =
  \phi(\mathbf{X})
  =
  \{\mathbf{z}_p\}_{p=1}^{N},
  \qquad
  \mathbf{z}_p \in \mathbb{R}^{d},
  \label{eq:patch_tokens}
\end{equation}
where $\mathbf{Z}$ denotes the set of all patch-level representations,
$\mathbf{z}_p$ is the $d$-dimensional semantic representation of the
$p$-th image patch, and $N$ is the number of image patches. The visual encoder is initialized from a pretrained model and adapted by the
gaze-ordered objective.

Each image is paired with a fixation sequence
$\mathcal{F}=\{f_1,\ldots,f_T\}$, where $f_t=(x_t,y_t)$ denotes the
coordinates of the $t$-th fixation. We assign each fixation to its nearest
image patch and merge repeated visits to the same patch while preserving the
first-visit order. This yields an ordered sequence of unique visited patch
indices, denoted by $\mathcal{S}$:
\begin{equation}
  \mathcal{S}
  =
  (s_1,\ldots,s_L),
  \qquad
  s_i \in \{1,\ldots,N\}.
  \label{eq:fixation_sequence}
\end{equation}

\subsection{Next-Fixation Prediction}
\label{sec:method:arjepa}

As illustrated in Fig.~\ref{fig:architecture}B, the next-fixation branch uses
an autoregressive predictor to forecast the latent representation of each
successive expert-fixated region from all preceding fixations. For each visited patch $s_i$ in $\mathcal{S}$, we construct a fixation token by combining its visual representation with positional and temporal cues:
\begin{equation}
  \mathbf{h}_i
  =
  \psi
  \left(
    \mathbf{z}_{s_i},
    \mathbf{e}_{s_i},
    i,
    \Delta_i
  \right),
  \label{eq:fixation_embedder}
\end{equation}
where $\mathbf{e}_{s_i}$ is the spatial embedding of patch $s_i$, $i$ the temporal rank of the fixation, $\Delta_i$ its dwell duration, and $\psi$ the fixation embedder. Following the JEPA~\citep{lecun2022path,assran2023self}, which learns by
predicting target representations rather than reconstructing raw inputs, an autoregressive predictor $g_{\theta}$ with a causal attention mask receives the prefix of visited fixation tokens and predicts the representation of the next fixated patch:
\begin{equation}
  \hat{\mathbf{z}}_{s_i}
  =
  g_{\theta}
  \left(
    \mathbf{h}_1,\ldots,\mathbf{h}_{i-1}
  \right),
  \qquad
  i=2,\ldots,L.
  \label{eq:ar_prediction}
\end{equation}
Prediction targets come from a momentum target encoder $\bar{\phi}$, an exponential-moving-average (EMA) copy of the online visual encoder $\phi$~\citep{grill2020bootstrap}. While $\phi$ is
directly optimized by gradient descent, $\bar{\phi}$ is updated via stop-gradient after each optimization step, providing a stable representation space for the prediction objective:
\begin{equation}
  \bar{\mathbf{z}}_{s_i}
  =
  \bar{\phi}(\mathbf{X})_{s_i},
  \label{eq:target_rep}
\end{equation}
\begin{equation}
  \bar{\theta}_{\phi}
  \leftarrow
  \tau \bar{\theta}_{\phi}
  +
  (1-\tau)\theta_{\phi},
  \label{eq:ema_update}
\end{equation}
where $\theta_{\phi}$ and $\bar{\theta}_{\phi}$ denote the parameters of the online visual encoder and the target encoder, respectively. The autoregressive loss is
\begin{equation}
  \mathcal{L}_{\mathrm{AR}}
  =
  \frac{1}{L-1}
  \sum_{i=2}^{L}
  \mathrm{SmoothL1}
  \left(
    \mathrm{LN}(\hat{\mathbf{z}}_{s_i}),
    \mathrm{sg}(\bar{\mathbf{z}}_{s_i})
  \right),
  \label{eq:loss_ar}
\end{equation}
where $\mathrm{LN}(\cdot)$ denotes layer normalization and $\mathrm{sg}(\cdot)$ stop-gradient. Following the JEPA principle, prediction operates in representation space rather than pixel space, encouraging the model to capture anatomical and diagnostic regularities that make expert fixation order predictable.
% ------------------------------------------------------------
\subsection{Spatial Completion}
\label{sec:method:completion}
% ------------------------------------------------------------

As shown in Fig.~\ref{fig:architecture}C, the spatial-completion branch
complements the autoregressive objective, which supervises only the patches
that appear in the radiologist's fixation sequence. However, unvisited regions can still be clinically informative. For example, a region may be skipped because it is normal, because it is peripheral to the suspected finding, or because the accumulated context is already sufficient. We therefore introduce a spatial-completion branch that assigns a representation-space target to every unvisited patch.

Let
\begin{equation}
  \mathcal{U}
  =
  \{1,\ldots,N\}
  \setminus
  \{s_1,\ldots,s_L\}
  \label{eq:unvisited_set}
\end{equation}
denote the set of unvisited patches. A lightweight cross-attention
decoder $d_{\xi}$ predicts one representation for each unvisited
patch. Queries combine a learnable mask token with the spatial embedding of the target patch; keys and values come from the gaze-context representation:
\begin{equation}
  \hat{\mathbf{r}}_p
  =
  d_{\xi}
  \left(
    \mathbf{m}_p + \mathbf{e}_{p},
    \mathbf{H}
  \right),
  \qquad
  p \in \mathcal{U},
  \label{eq:completion_prediction}
\end{equation}
where
\begin{equation}
  \mathbf{H}
  =
  g_{\theta}
  \left(
    \mathbf{h}_1,\ldots,\mathbf{h}_{L}
  \right)
  \label{eq:gaze_context}
\end{equation}
is the full fixation-context representation produced by the autoregressive predictor. The spatial-completion loss matches each prediction to the corresponding momentum target:
\begin{equation}
  \mathcal{L}_{\mathrm{SC}}
  =
  \frac{1}{|\mathcal{U}|}
  \sum_{p \in \mathcal{U}}
  \mathrm{SmoothL1}
  \left(
    \mathrm{LN}(\hat{\mathbf{r}}_{p}),
    \mathrm{sg}
    \left(
      \mathrm{LN}(\bar{\mathbf{z}}_{p})
    \right)
  \right).
  \label{eq:loss_sc}
\end{equation}

This objective resembles masked representation prediction~\citep{assran2023self}, but derives its context from the radiologist's reading trajectory rather than random visible patches. The model thus learns to infer unvisited anatomy from the evidence that an expert chose to inspect.

% ------------------------------------------------------------
\subsection{Pretraining Objective}
\label{sec:method:objective}
% ------------------------------------------------------------

The complete world-model pretraining objective combines
autoregressive next-fixation prediction and spatial completion:
\begin{equation}
  \mathcal{L}_{\mathrm{pre}}
  =
  \mathcal{L}_{\mathrm{AR}}
  +
  \lambda
  \mathcal{L}_{\mathrm{SC}}.
  \label{eq:pretrain_loss}
\end{equation}

During pretraining, the online visual encoder is optimized jointly with the fixation embedder, autoregressive predictor, prediction head, and spatial-completion decoder. The target encoder receives no gradients and is updated only through EMA~\eqref{eq:ema_update}. 

%% file: experiments.tex
% ============================================================
%  §4  Experiments
% ============================================================

\section{Experiments}
\label{sec:experiments}

% ------------------------------------------------------------
\subsection{Experimental Setup}
\label{sec:experiments:setup}
% ------------------------------------------------------------

We evaluate GazeWorld on diagnostic classification and scanpath prediction.

\paragraph{Diagnostic classification.}
We evaluate frozen backbone features of GazeWorld on CheXpert~\citep{irvin2019chexpert}, RSNA Pneumonia~\citep{colak2021rsna}, and SIIM-ACR Pneumothorax~\citep{siim_acr_pneumothorax_2019} under $1\%$, $10\%$, and $100\%$ label fractions, as well as zero-shot transfer on all three benchmarks. We compare against BioViL~\citep{bannur2023learning}, MedKLIP~\citep{wu2023medklip}, MGCA~\citep{wang2022multi}, GLoRIA~\citep{huang2021gloria}, PRIOR~\citep{cheng2023prior}, MedCLIP~\citep{wang2022medclip}, CheXWorld~\citep{yue2025chexworld}, and EGMA~\citep{ma2024eye}. 

\paragraph{Scanpath prediction.}
We pretrain on MIMIC-EYE~\citep{hsieh2023mimic} and probe on GazeSearch~\citep{pham2025gazesearch}, a refined scanpath benchmark derived from MIMIC-EYE with more precise gaze-path annotations, under the ChestSearch protocol. For each backbone variant, we freeze the pretrained visual encoder and train only the same lightweight
scanpath decoder on top of its patch-level features; performance differences therefore reflect scanpath-relevant information already encoded in the frozen representations. We compare against representative scanpath prediction methods, including
general visual-search baselines GazeFormer~\citep{mondal2023gazeformer}
and HAT~\citep{yang2024unifying}, radiology-specific GazeSearch
~\citep{pham2025gazesearch}, and the recent LogitGaze /
LogitGaze-Med models~\citep{lvovhuman}, with LogitGaze-Med serving as the
strongest scanpath baseline.

All dataset statistics, baseline protocols, and split details are provided in Appendices~\ref{sec:appendix:datasets} and~\ref{sec:appendix:baseline_protocol}.

% We evaluate GazeWorld along two axes. First, we test whether
% world-model pretraining on fixation sequences produces representations
% that preserve the sequential structure of expert reading. For this
% purpose, we use MIMIC-EYE~\citep{hsieh2023mimic} for pretraining and
% GazeSearch~\citep{pham2025gazesearch}, a refined scanpath benchmark
% derived from MIMIC-EYE with more precise gaze-path annotations, for
% probing under the ChestSearch protocol. For each backbone variant, we
% freeze the pretrained visual encoder and train only the same
% lightweight scanpath decoder on top of its patch-level features
% (Appendix~\ref{sec:appendix:scanpath}); thus, performance differences
% reflect scanpath-relevant information already encoded in the frozen
% representations.

% Second, we test whether these representations transfer to diagnostic
% classification. We evaluate frozen backbone features on CheXpert,
% RSNA Pneumonia, and SIIM-ACR Pneumothorax under $1\%$, $10\%$, and
% $100\%$ label fractions, as well as zero-shot transfer on the same
% external benchmarks. Dataset details and split statistics are provided
% in Appendix~\ref{sec:appendix:datasets}.

% For scanpath probing, we compare against GazeFormer, HAT,
% GazeSearch, LogitGaze, and LogitGaze-Med. For classification, we
% compare against BioViL, MedKLIP, MGCA, GLoRIA, PRIOR, MedCLIP,
% CheXWorld, and EGMA. We follow the unified baseline protocol
% summarized in Appendix~\ref{sec:appendix:baseline_protocol}.

\subsection{Implementation Details}
We use MedCLIP-Swin~\citep{liu2021swin} as the visual encoder,
processing chest radiographs at $224 \times 224$ resolution. The
autoregressive predictor is an 8-layer causal Transformer with
hidden dimension $768$. GazeWorld is pretrained on
MIMIC-EYE~\citep{hsieh2023mimic} for 15 epochs using
AdamW~\citep{loshchilov2017decoupled} with learning rate
$3 \times 10^{-4}$. The spatial-completion weight is set to
$\lambda = 1.0$. At evaluation, the visual encoder and predictor
are frozen; downstream tasks use only the resulting representations
with no access to gaze data. Full architecture specifications,
hyperparameters, and evaluation protocols are provided
in Appendix~\ref{sec:appendix:impl}.

% ------------------------------------------------------------

% ------------------------------------------------------------
\subsection{Downstream Classification}
\label{sec:experiments:cls}
% ------------------------------------------------------------

If the world model has captured the invariants of expert reading, its
representations should also transfer to standard diagnostic tasks. We
evaluate frozen backbone features for supervised and zero-shot
classification on three external datasets that are not used during
world-model pretraining.

\begin{table*}[t]
  \centering
  \caption{
    Comparison of supervised classification AUROC on
    CheXpert~\citep{irvin2019chexpert}, RSNA~\citep{colak2021rsna},
    and SIIM-ACR~\citep{siim_acr_pneumothorax_2019} with $1\%$,
    $10\%$, and $100\%$ of training labels.
    \best{Best} and \second{second-best} results are highlighted.
  }
  \label{tab:supervised}
  \small
  \setlength{\tabcolsep}{4.5pt}
  \renewcommand{\arraystretch}{1.15}
  \begin{tabular}{l*{9}{c}}
    \toprule
    \multirow{2}{*}{Method}
      & \multicolumn{3}{c}{\textbf{CheXpert}}
      & \multicolumn{3}{c}{\textbf{RSNA}}
      & \multicolumn{3}{c}{\textbf{SIIM-ACR}} \\
    \cmidrule(lr){2-4} \cmidrule(lr){5-7} \cmidrule(lr){8-10}
      & 1\% & 10\% & 100\%
      & 1\% & 10\% & 100\%
      & 1\% & 10\% & 100\% \\
    \midrule
    BioViL~\citep{bannur2023learning}
      & 54.19 & 60.37 & 66.55
      & 70.74 & 72.84 & 78.48
      & 62.98 & 68.39 & 78.20 \\
    MedKLIP~\citep{wu2023medklip}
      & -     & -     & -
      & 79.43 & 78.62 & 82.76
      & 74.50 & 78.27 & 84.67 \\
    MGCA~\citep{wang2022multi}
      & 67.12 & 76.10 & 81.57
      & 80.41 & 85.67 & 88.24
      & 86.14 & \second{88.13} & \second{91.84} \\
    GLoRIA~\citep{huang2021gloria}
      & 67.40 & 77.37 & 81.74
      & 81.90 & 81.24 & 88.57
      & -     & -     & -     \\
    PRIOR~\citep{cheng2023prior}
      & 65.51 & 70.70 & 79.11
      & 79.74 & 79.72 & 83.95
      & 81.96 & 82.76 & 85.33 \\
    CheXWorld~\citep{yue2025chexworld}
      & 62.81 & 80.03 & 85.39
      & 78.30 & 84.12 & \second{88.66}
      & 62.31 & 81.54 & 91.34 \\
    MedCLIP~\citep{wang2022medclip}
      & \second{74.65} & \second{82.88} & \second{85.58}
      & 82.31 & 84.16 & 87.86
      & 86.09 & 87.41 & 91.46 \\
    EGMA~\citep{ma2024eye}
      & 71.21 & 79.79 & 85.45
      & \second{82.49} & \second{84.73} & 88.09
      & \second{87.17} & 87.91 & 88.98 \\
    Ours
      & \best{78.37} & \best{83.41} & \best{87.61}
      & \best{83.27} & \best{85.75} & \best{90.15}
      & \best{87.85} & \best{90.48} & \best{94.27} \\
    \bottomrule
  \end{tabular}
\end{table*}
\begin{figure*}[h]
  \centering
  \includegraphics[width=\textwidth]{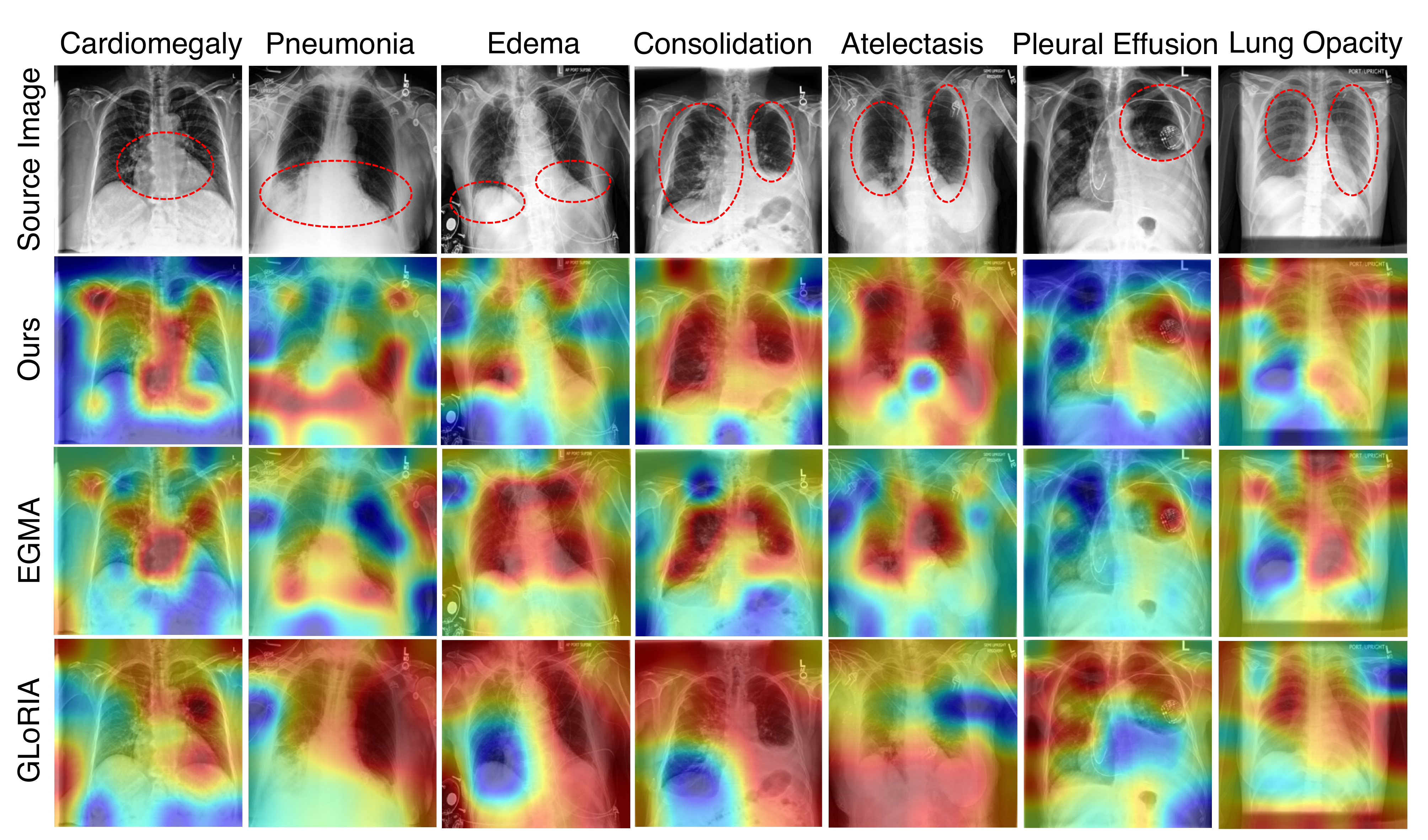}
  \caption{%
    Grad-CAM~\citep{selvaraju2017grad} attention visualizations across
    seven pathologies, comparing the source radiograph, GazeWorld,
    EGMA~\citep{ma2024eye}, and GLoRIA~\citep{huang2021gloria}.
    \textcolor{red}{Red dashed circles} on the source radiographs
    highlight pathology-relevant regions used for visual reference.
    All maps are generated with the same Grad-CAM protocol
    (Appendix~\ref{sec:appendix:heatmap}).%
  }
  \label{fig:heatmap}
\end{figure*}
\paragraph{Classification results.}
Tables~\ref{tab:supervised} and~\ref{tab:zeroshot} summarize
supervised and zero-shot diagnostic transfer. In supervised
classification, GazeWorld achieves the best AUROC in all nine
dataset--label settings. Under the controlled comparison with EGMA,
where the Swin backbone, MedCLIP initialization, gaze data, and
downstream protocol are matched, GazeWorld improves CheXpert $1\%$
AUROC from $71.21$ to $78.37$. It also outperforms
CheXWorld~\citep{yue2025chexworld} in all settings, with the
largest gains in low-label regimes. In zero-shot evaluation,
GazeWorld obtains the highest accuracy on all three datasets and the
highest F1 on CheXpert and SIIM-ACR. Figure~\ref{fig:heatmap} provides qualitative Grad-CAM visualizations
under the EGMA protocol (Appendix~\ref{sec:appendix:heatmap}), showing
that GazeWorld produces more concentrated activations over clinically
relevant anatomy than EGMA and GLoRIA. These visualizations are
qualitative and are not intended as formal localization evaluations.
Training and evaluation details are summarized in Appendix~\ref{sec:appendix:impl}.

\begin{table}[t]
  \centering
  \caption{
    Zero-shot classification on
    CheXpert 5$\times$200~\citep{irvin2019chexpert},
    RSNA~\citep{colak2021rsna}, and
    SIIM-ACR~\citep{siim_acr_pneumothorax_2019}.
    AUROC, Accuracy, and F1 are computed without task-specific
    fine-tuning. \best{Best} and \second{second-best} results are highlighted.
  }
  \label{tab:zeroshot}
  \small
  \renewcommand{\arraystretch}{1.08}
  \setlength{\tabcolsep}{3pt}
  \resizebox{\linewidth}{!}{
  \begin{tabular}{lcccccc}
    \toprule
    \multirow{2}{*}{\textbf{Method}}
      & \multicolumn{2}{c}{\textbf{CheXpert 5$\times$200}}
      & \multicolumn{2}{c}{\textbf{RSNA}}
      & \multicolumn{2}{c}{\textbf{SIIM-ACR}} \\
    \cmidrule(lr){2-3} \cmidrule(lr){4-5} \cmidrule(lr){6-7}
      & \textbf{AUROC}$\uparrow$ & \textbf{Acc.$\uparrow$ / F1}$\uparrow$
      & \textbf{AUROC}$\uparrow$ & \textbf{Acc.$\uparrow$ / F1}$\uparrow$
      & \textbf{AUROC}$\uparrow$ & \textbf{Acc.$\uparrow$ / F1}$\uparrow$ \\
    \midrule
    CLIP~\citep{radford2021learning}
      & 15.50 & 19.60 / 10.99
      & 21.67 & 23.46 / 19.87
      & 42.13 & 46.11 / 40.13 \\
    GLoRIA~\citep{huang2021gloria}
      & 49.18 & 51.10 / 47.25
      & \second{60.77} & 61.53 / \best{60.09}
      & -- & - / - \\
    MGCA~\citep{wang2022multi}
      & 42.53 & 43.39 / 39.67
      & 59.19 & 60.82 / \second{59.57}
      & 36.60 & 38.45 / 34.95 \\
    MedCLIP~\citep{wang2022medclip}
      & \second{56.63} & \second{57.40} / \second{55.72}
      & 58.08 & 62.17 / 40.06
      & 56.17 & 56.89 / 55.44 \\
    EGMA~\citep{ma2024eye}
      & 56.09 & 57.15 / 55.03
      & 57.46 & \second{66.85} / 43.08
      & \second{59.79} & \second{63.24} / \second{61.04} \\
    Ours
      & \best{58.13} & \best{59.42} / \best{56.85}
      & \best{62.84} & \best{70.18} / 47.31
      & \best{64.29} & \best{66.59} / \best{63.26} \\
    \bottomrule
  \end{tabular}
  }
\end{table}

% \paragraph{Zero-shot classification.}
% Table~\ref{tab:zeroshot} reports zero-shot transfer. GazeWorld
% achieves the highest accuracy on all three datasets and the highest
% F1 on CheXpert and SIIM-ACR. The consistent advantage over EGMA under
% the same backbone family further supports the contribution of
% gaze-ordered world-model pretraining.

% Figure~\ref{fig:heatmap} provides qualitative Grad-CAM
% visualizations under the EGMA protocol
% (Appendix~\ref{sec:appendix:heatmap}). Compared with EGMA and GLoRIA,
% GazeWorld produces more concentrated activations over clinically
% relevant anatomy, such as the cardiac silhouette, costophrenic sulcus,
% and focal opacities. These visualizations are qualitative and are not
% intended as formal localization evaluation.
\begin{figure*}[h]
  \centering
  \includegraphics[width=\textwidth]{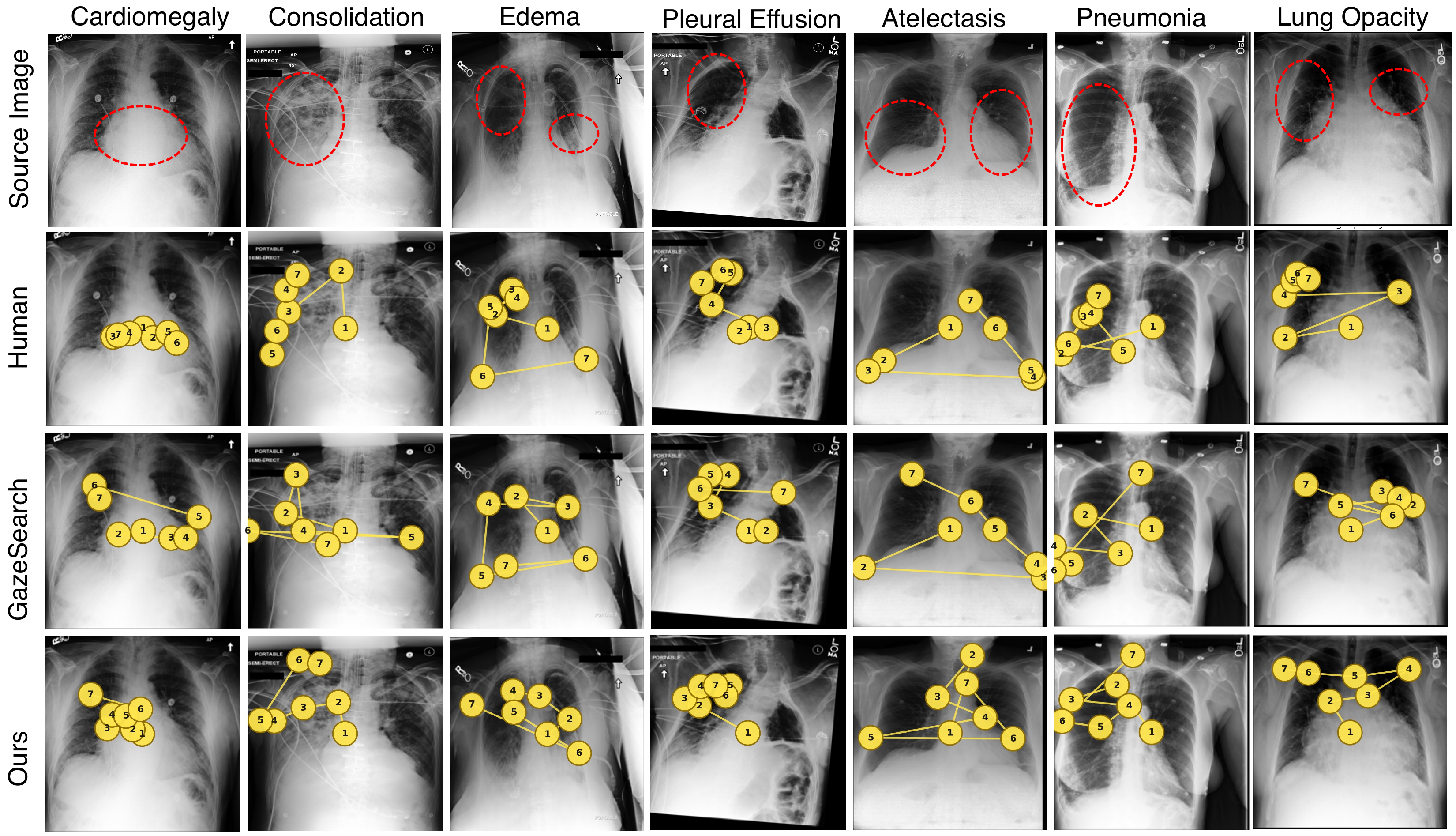}
  \caption{%
    Qualitative scanpath comparison across seven pathologies.
    The first row shows the source radiographs, with
    \textcolor{red}{red dashed circles} marking pathology-relevant
    regions. We compare human scanpaths, GazeSearch
    predictions~\citep{pham2025gazesearch}, and trajectories decoded
    from frozen GazeWorld representations using the same supervised
    decoder.%
  }
  \label{fig:scanpath}
\end{figure*}

\subsection{Scanpath Prediction}
\label{sec:experiments:scanpath}
% ------------------------------------------------------------

A central claim of this work is that world-model pretraining on
fixation sequences produces representations that encode the sequential
structure of expert reading. We test this claim with a controlled
probing experiment. For each backbone variant, we freeze the
pretrained visual encoder and train only a lightweight scanpath
decoder on top of its patch-level features
(Appendix~\ref{sec:appendix:scanpath}). The decoder architecture,
optimization settings, and training data are kept identical across
variants; the only variable is the frozen representation. Thus,
performance differences reflect how much scanpath-relevant information
is already encoded in the learned features.

\begin{table*}[t]
  \centering
  \caption{
    GazeSearch scanpath probing results. ScanMatch, STDE, and MultiMatch
    scores are higher-better, while SED is lower-better. \best{Best} and
    \second{second-best} results are highlighted.
  }
  \label{tab:scanpath_similarity_multimatch}
  \footnotesize
  \setlength{\tabcolsep}{4pt}
  \renewcommand{\arraystretch}{1.10}
  \resizebox{\textwidth}{!}{%
  \begin{tabular}{lcccccc}
    \toprule
    \multirow{2}{*}{\textbf{Method}}
      & \multicolumn{3}{c}{\textbf{Scanpath}}
      & \multicolumn{3}{c}{\textbf{MultiMatch}} \\
    \cmidrule(lr){2-4}
    \cmidrule(lr){5-7}
      & \textbf{SM}$\uparrow$
      & \textbf{SED}$\downarrow$
      & \textbf{STDE}$\uparrow$
      & \textbf{Vec.}$\uparrow$
      & \textbf{Dir.}$\uparrow$
      & \textbf{Pos.}$\uparrow$ \\
    \midrule
    GazeFormer~\citep{mondal2023gazeformer}
      & 0.280 & 5.11 & 0.801
      & 0.902 & 0.644 & 0.803 \\
    HAT~\citep{yang2024unifying}
      & 0.296 & 5.07 & 0.800
      & 0.909 & 0.649 & 0.825 \\
    GazeSearch~\citep{pham2025gazesearch}
      & 0.351 & 4.89 & 0.809
      & 0.917 & 0.679 & 0.829 \\
    LogitGaze~\citep{lvovhuman}
      & 0.328 & 5.07 & 0.810
      & 0.882 & 0.643 & 0.809 \\
    LogitGaze-Med (Res)~\citep{lvovhuman}
      & 0.416 & 4.68 & 0.852
      & 0.935 & 0.650 & 0.823 \\
    LogitGaze-Med (CheX)~\citep{lvovhuman}
      & 0.419 & 4.68 & 0.855
      & 0.938 & 0.651 & 0.823 \\
    \midrule
    Ours (MedCLIP)
      & 0.479 & 3.74 & 0.909
      & 0.941 & 0.649 & 0.837 \\
    Ours (Res)
      & \second{0.484} & \second{3.66} & \best{0.912}
      & \second{0.966} & \second{0.678} & \best{0.917} \\
    Ours (CheX)
      & \best{0.489} & \best{3.63} & \second{0.911}
      & \best{0.966} & \best{0.686} & \second{0.915} \\
    \bottomrule
  \end{tabular}
  }
\end{table*}

Table~\ref{tab:scanpath_similarity_multimatch} shows that all three variants of
GazeWorld outperform the previous state-of-the-art LogitGaze-Med on the main
scanpath metrics. The best variant improves
ScanMatch~\citep{cristino2010scanmatch} by over $16\%$, reduces
SED~\citep{foulsham2008can} by more than $22\%$, and raises
STDE~\citep{wang2011simulating} by over $6\%$ relative to
LogitGaze-Med~(CheX). Since LogitGaze-Med is a purpose-built scanpath
predictor with explicit gaze supervision, whereas our probing uses the same
generic decoder on frozen features, these gains suggest that gaze-ordered
world-model pretraining encodes scanpath-relevant structure into the learned
representation. The same table further reports
MultiMatch~\citep{jarodzka2010vector} results, with consistent gains across vector, direction, and position.

Figure~\ref{fig:scanpath} provides qualitative support for the
probing results. Across seven pathologies, scanpaths decoded from
GazeWorld features more closely follow radiologist reading
trajectories and attend to pathology-relevant anatomy, whereas
GazeSearch produces more diffuse fixation sequences.
% ------------------------------------------------------------
\subsection{Ablation Studies}
\label{sec:experiments:ablation}
% ------------------------------------------------------------
\paragraph{Gaze ordering and objectives.}
Table~\ref{tab:ablation} shows that chronological gaze order is
critical: replacing expert scanpaths with raster-scan or random order
substantially degrades zero-shot AUROC across all datasets, indicating
that the gain is not merely due to ordered patch exposure. MedCLIP
initialization performs best, suggesting that gaze-ordered prediction
benefits from a semantically aligned medical visual backbone. Finally,
AR-only already gives strong performance, while adding spatial
completion further improves results; SC-only is weaker, suggesting that predicting unvisited anatomy without the
autoregressive fixation context provides less effective supervision, and that completion is most beneficial when conditioned on chronological expert
fixation dynamics.

\paragraph{Backbone initialization.}
Table~\ref{tab:init_ablation_compact} further evaluates whether GazeWorld can
improve different medical visual backbones. Across MGCA, EGMA, and MedCLIP
initializations, adding GazeWorld consistently improves zero-shot performance
on the external benchmarks, showing that the proposed gaze-ordered world-model
pretraining is not tied to a single backbone. These results suggest that GazeWorld can generally refine medical visual encoders with expert reading dynamics.

% We ablate three design factors: fixation ordering, backbone
% initialization, and the two components of the proposed pretraining
% objective. All metrics in Table~\ref{tab:ablation} are zero-shot,
% following the EGMA evaluation protocol~\citep{ma2024eye}.

\begin{table}[t]
  \centering
  \caption{
    Ablation on fixation ordering, backbone initialization, and
    objective components. ``AR only'' denotes next-fixation prediction
    without spatial completion, and ``SC only'' denotes spatial
    completion without autoregressive prediction. All metrics are
    \textbf{zero-shot}: AUROC, Accuracy, and F1 are computed without
    task-specific fine-tuning.
  }
  \label{tab:ablation}
  \small
  \renewcommand{\arraystretch}{1.15}
  \begin{tabular}{lcccccc}
    \toprule
    \multirow{2}{*}{\textbf{Variant}}
      & \multicolumn{2}{c}{\textbf{CheXpert 5$\times$200}}
      & \multicolumn{2}{c}{\textbf{RSNA}}
      & \multicolumn{2}{c}{\textbf{SIIM-ACR}} \\
    \cmidrule(lr){2-3} \cmidrule(lr){4-5} \cmidrule(lr){6-7}
      & \textbf{AUROC}$\uparrow$ & \textbf{Acc.}$\uparrow$ / \textbf{F1}$\uparrow$
      & \textbf{AUROC}$\uparrow$ & \textbf{Acc.}$\uparrow$ / \textbf{F1}$\uparrow$
      & \textbf{AUROC}$\uparrow$ & \textbf{Acc.}$\uparrow$ / \textbf{F1}$\uparrow$ \\
    \midrule
    Raster-scan Order
      & 52.18 & 55.21 / 51.58
      & 50.39 & 62.46 / 42.30
      & 53.50 & 56.22 / 50.16 \\
    Random Order
      & 52.25 & 56.08 / 49.13
      & 51.58 & 61.53 / 42.68
      & 55.06 & 57.02 / 54.34 \\
    \midrule
    % MGCA init
    %   & 47.41 & 48.31 / 46.07
    %   & 57.76 & 61.46 / 59.98
    %   & 41.07 & 43.56 / 39.62 \\
    % EGMA init
    %   & 55.12 & 56.30 / 54.45
    %   & 58.50 & 67.16 / 44.89
    %   & 60.83 & 64.01 / 61.33 \\
    % \midrule
    AR Only
      & 57.78 & 58.86 / 55.91
      & 59.11 & 68.07 / 44.15
      & 62.10 & 64.88 / 61.94 \\
    SC Only
      & 54.67 & 54.41 / 53.06
      & 51.25 & 60.63 / 40.03
      & 52.59 & 53.72 / 51.37 \\
    \midrule
    Ours
      & 58.13 & 59.42 / 56.85
      & 62.84 & 70.18 / 47.31
      & 64.29 & 66.59 / 63.26 \\
    \bottomrule
  \end{tabular}
\end{table}

\begin{table*}[t]
  \centering
  \caption{
    Effect of backbone initialization on zero-shot classification.
    AUROC, Accuracy, and F1 are reported on three external datasets.
  }
  \label{tab:init_ablation_compact}
  \small
  \setlength{\tabcolsep}{4pt}
  \renewcommand{\arraystretch}{1.12}
  \begin{tabular}{lccccccccc}
    \toprule
    \multirow{2}{*}{\textbf{Backbone}}
      & \multicolumn{3}{c}{\textbf{CheXpert 5$\times$200}}
      & \multicolumn{3}{c}{\textbf{RSNA}}
      & \multicolumn{3}{c}{\textbf{SIIM-ACR}} \\
    \cmidrule(lr){2-4}
    \cmidrule(lr){5-7}
    \cmidrule(lr){8-10}
      & \textbf{AUROC}$\uparrow$ & \textbf{Acc.}$\uparrow$ & \textbf{F1}$\uparrow$
      & \textbf{AUROC}$\uparrow$ & \textbf{Acc.}$\uparrow$ & \textbf{F1}$\uparrow$
      & \textbf{AUROC}$\uparrow$ & \textbf{Acc.}$\uparrow$ & \textbf{F1}$\uparrow$ \\
    \midrule
    MGCA
      & 42.53 & 43.39 & 39.67
      & 59.19 & 60.82 & 59.57
      & 36.60 & 38.45 & 34.95 \\
    \textbf{+GazeWorld}
      & 47.41 & 48.31 & 46.07
      & 60.76 & 61.46 & 59.98
      & 41.07 & 43.56 & 39.62 \\
    $\Delta$
      & \textcolor{green!50!black}{+4.88} & \textcolor{green!50!black}{+4.92} & \textcolor{green!50!black}{+6.40}
      & +1.57 & +0.64 & +0.41
      & +4.47 & +5.11 & +4.67 \\
    \midrule
    EGMA
      & 56.09 & 57.15 & 55.03
      & 57.46 & 66.85 & 43.08
      & 59.77 & 63.24 & 61.04 \\
    \textbf{+GazeWorld}
      & 56.42 & 58.30 & 55.45
      & 58.50 & 67.16 & 44.89
      & 60.83 & 64.01 & 61.33 \\
    $\Delta$
      & +0.33 & +1.15 & +0.42
      & +1.04 & +0.31 & +1.81
      & +1.06 & +0.77 & +0.29 \\
    \midrule
    MedCLIP
      & 56.63 & 57.40 & 55.72
      & 58.09 & 62.17 & 40.07
      & 56.17 & 56.89 & 55.44 \\
    \textbf{+GazeWorld}
      & 58.13 & 59.42 & 56.85
      & 62.84 & 70.18 & 47.31
      & 64.29 & 66.59 & 63.26 \\
    $\Delta$
      & +1.50 & +2.02 & +1.13
      & \textcolor{green!50!black}{+4.75} & \textcolor{green!50!black}{+8.01} & \textcolor{green!50!black}{+7.24}
      & \textcolor{green!50!black}{+8.12} & \textcolor{green!50!black}{+9.70} & \textcolor{green!50!black}{+7.82} \\
    \bottomrule
  \end{tabular}
\end{table*}

% \paragraph{Effect of fixation ordering.}
% Table~\ref{tab:ablation} shows that the chronological structure of
% expert gaze is a key source of supervision. Replacing the radiologist
% scanpath with raster-scan or random order causes zero-shot AUROC to
% drop toward chance across all three benchmarks. The similar
% performance of raster and random ordering indicates that the gain does
% not come merely from exposing the model to an ordered sequence of
% patches; rather, the temporal order imposed by expert reading carries
% diagnostic structure.

% \paragraph{Effect of backbone initialization.}
% Table~\ref{tab:ablation} shows that the world-model objective depends
% on the semantic quality of the initial visual encoder. MedCLIP
% initialization gives the best zero-shot AUROC on CheXpert, RSNA, and
% SIIM-ACR, while EGMA provides the second-best initialization and MGCA
% is substantially weaker. These results support applying gaze-ordered
% predictive learning on top of a medical vision-language backbone.

% \paragraph{Effect of objective components.}
% The bottom block of Table~\ref{tab:ablation} decomposes the objective
% into autoregressive next-fixation prediction and spatial completion.
% AR-only already performs strongly, confirming gaze-ordered prediction
% as the main learning signal. Spatial completion provides complementary
% gains by supervising unvisited patches, while SC-only remains weaker,
% showing that completion is most effective when paired with
% chronological fixation modeling.

%% file: conclusion.tex
% ============================================================
%  §5  Conclusion and Limitations
% ============================================================
\section{Conclusion}
\label{sec:conclusion}

We introduced GazeWorld, a gaze-ordered world-model refinement objective
for medical image representation learning. By predicting the next
expert-fixated representation and completing unvisited anatomy in semantic
space, GazeWorld injects radiologist reading dynamics into pretrained
medical visual encoders. Experiments on diagnostic transfer, scanpath
probing, and backbone initialization show that this objective improves
task performance and better encodes expert visual-search structure,
suggesting that modeling how radiologists gather evidence can complement
existing medical pretraining paradigms. 

\paragraph{Limitations}. This study is limited to chest radiographs and one eye-tracking corpus.
Future work should evaluate larger multi-reader gaze datasets, broader
imaging modalities, and formal clinician-centered localization studies.

%% file: appendix.tex
% ============================================================
%  Appendix
% ============================================================

\appendix

% \section{Additional Scanpath Metrics}
% \label{sec:appendix:multimatch}

% Table~\ref{tab:multimatch} reports MultiMatch
% similarity~\citep{jarodzka2010vector}, which decomposes
% scanpath comparison into vector (saccade amplitude), direction
% (saccade angle), and position (fixation location) components.
% GazeWorld achieves the highest scores on all three
% components, with especially large gains on position similarity
% ($0.917$ vs.\ $0.823$ for LogitGaze-Med), confirming that the
% world-model representations encode fine-grained spatial structure
% of expert reading beyond what aggregate metrics capture.

% \begin{table}[h]
%   \centering
%   \caption{
%     MultiMatch similarity (higher is better) across three components.
%   }
%   \label{tab:multimatch}
%   \small
%   \begin{tabular}{lccc}
%     \toprule
%     \textbf{Method}
%       & \textbf{Vector} $\uparrow$
%       & \textbf{Direction} $\uparrow$
%       & \textbf{Position} $\uparrow$ \\
%     \midrule
%     GazeFormer
%       & 0.902 & 0.644 & 0.803 \\
%     HAT
%       & 0.909 & 0.649 & 0.825 \\
%     GazeSearch
%       & 0.917 & 0.679 & 0.829 \\
%     LogitGaze
%       & 0.882 & 0.643 & 0.809 \\
%     LogitGaze-Med (Res)
%       & 0.935 & 0.650 & 0.823 \\
%     LogitGaze-Med (CheX)
%       & 0.938 & 0.651 & 0.823 \\
%     \midrule
%     Ours (Res)
%       & 0.966 & 0.678 & 0.917 \\
%     Ours (CheX)
%       & 0.966 & 0.686 & 0.915 \\
%     \bottomrule
%   \end{tabular}
% \end{table}
\section{Baseline Protocol Details}
\label{sec:appendix:baselines}

\subsection{Dataset Details}
\label{sec:appendix:datasets}

\paragraph{MIMIC-EYE.}
MIMIC-EYE~\citep{hsieh2023mimic} is an eye-tracking
extension of the MIMIC-CXR database~\citep{johnson2019mimic}.
It records radiologist fixation sequences from $3{,}032$ frontal
chest radiographs during routine clinical reading sessions using a
Tobii Pro Nano eye tracker (sampling rate 60\,Hz).
Each record contains a temporally ordered sequence of fixation
coordinates $(x_t, y_t)$ with associated dwell durations.
We split the data into $2{,}729$ / $151$ / $152$
train/val/test following the official partition.
Per-image fixation sequences average $47$ fixation points; after
de-duplication to unique patch indices, the resulting sequence is
used as the causal order for world-model pretraining.

\paragraph{GazeSearch (ChestSearch protocol).} GazeSearch~\citep{pham2025gazesearch} provides task-directed
eye-tracking data: radiologists search for specific pathologies
(13 finding categories) on chest radiographs.
The ChestSearch split contains $3{,}870$ training and $488$ test
samples, each consisting of a radiograph, a target pathology label,
and a ground-truth fixation sequence (up to 7 fixations).

\paragraph{CheXpert.}
CheXpert~\citep{irvin2019chexpert} is a large-scale chest
radiograph dataset containing $224{,}316$ images from $65{,}240$
patients at Stanford Hospital, labeled for 14 observations via an
automated NLP labeler applied to radiology reports.
We evaluate on 5 competition pathologies (Atelectasis, Cardiomegaly,
Consolidation, Edema, Pleural Effusion) under $1\%$, $10\%$, and
$100\%$ label fractions.
For zero-shot evaluation, we follow the CheXpert 5$\times$200
protocol~\citep{ma2024eye}: 200 images per class, 5 pathologies.

\paragraph{RSNA Pneumonia.}
The RSNA~\citep{colak2021rsna} Pneumonia Detection Challenge dataset provides $26{,}684$
frontal chest radiographs with bounding-box annotations for
pneumonia opacities, curated from the NIH ChestX-ray14 collection.
We use the binary pneumonia/normal classification task under
$1\%$, $10\%$, and $100\%$ label fractions.

\paragraph{SIIM-ACR Pneumothorax.}
The SIIM-ACR Pneumothorax Segmentation dataset~\citep{siim_acr_pneumothorax_2019} contains $12{,}047$
chest radiographs with pixel-level pneumothorax masks.
We convert it to a binary classification task
(pneumothorax-present vs. normal) and evaluate under $1\%$,
$10\%$, and $100\%$ label fractions.

\subsection{Baseline Comparison Protocol}
\label{sec:appendix:baseline_protocol}

Table~\ref{tab:baseline_protocol} summarizes the backbone
architecture, pretraining data, and result provenance for every
baseline in the main comparison tables.

\paragraph{Classification baselines.}
All classification numbers for BioViL, MedKLIP, MGCA, GLoRIA, PRIOR,
MedCLIP, and EGMA in Tables~\ref{tab:supervised}
and~\ref{tab:zeroshot} are taken from EGMA~\citep{ma2024eye}, which
re-evaluated every method under a single downstream protocol.
The EGMA protocol fixes the following variables across all methods:
(i)~train/val/test splits,
(ii)~downstream head,
(iii)~optimizer and schedule,
(iv)~number of fine-tuning epochs, and early-stopping criterion, and
(v)~metric computation.
The protocol does not harmonize backbone architecture: BioViL,
MedKLIP, MGCA, GLoRIA, and PRIOR use ResNet-50, while MedCLIP, EGMA,
and our method use Swin-Tiny. Therefore, comparisons against
ResNet-50-based baselines include both objective and backbone
differences.

\paragraph{Controlled comparison (Ours vs.\ EGMA).}
The cleanest comparison in our evaluation is Ours vs.\ EGMA.
Both methods share the same Swin backbone, the same MedCLIP
initialization weights, the same MIMIC-EYE fixation data, and the
same downstream evaluation protocol.
The only variable is the pretraining objective applied on top
of MedCLIP: EGMA uses gaze-guided contrastive alignment, while we use
the GazeWorld world-model objective.
Performance differences between these two rows in
Tables~\ref{tab:supervised}--\ref{tab:zeroshot} therefore isolate
the contribution of the pretraining objective with minimal confounds.

\paragraph{Scanpath baselines.}
Scanpath prediction baselines in Table~\ref{tab:scanpath_similarity_multimatch}
are evaluated under the ChestSearch protocol. We use the original
paper results for LogitGaze-Med and evaluate GazeFormer, HAT,
GazeSearch, and LogitGaze using their official implementations under
the same protocol.
Our scanpath decoder uses the same evaluation protocol
(7 fixations, the same test split of 488 samples, and the same metrics).
For the scanpath probe, the decoder architecture and training
procedure are held fixed across backbone variants.

\paragraph{Resolution and backbone notes.}
For classification (Tables~\ref{tab:supervised}
and~\ref{tab:zeroshot}), all methods use $224 \times 224$ input
resolution. For scanpath prediction
(Table~\ref{tab:scanpath_similarity_multimatch}), our decoder and the primary
baselines operate at $448 \times 448$; GazeFormer and HAT use
$224 \times 224$ but are included for completeness as earlier methods.

% ----- Implementation Details -----
\section{Implementation Details}
\label{sec:appendix:impl}

\paragraph{Backbone.}
We use the MedCLIP Swin image encoder~\citep{liu2021swin}
as the online visual backbone. The encoder has 4 stages, embedding
dimension $C = 96$, window size 7, patch size 4, and output dimension
$d = 768$. Unless otherwise specified, chest radiographs are resized
to $224 \times 224$ for representation learning, yielding
patch-level semantic tokens
\begin{equation}
  \mathbf{Z}
  =
  \phi(\mathbf{X})
  =
  \{\mathbf{z}_p\}_{p=1}^{N},
  \qquad
  \mathbf{z}_p \in \mathbb{R}^{768}.
\end{equation}
During world-model pretraining, the online visual encoder is updated
jointly with the gaze-conditioned modules. A separate momentum target
encoder provides stop-gradient representation-space targets.

\paragraph{Fixation preprocessing.}
Each eye-tracking record provides a temporally ordered sequence of
fixation coordinates with associated dwell durations. Fixation
coordinates are normalized to the image frame and mapped to the
nearest patch index. Repeated visits to the same patch are merged
while preserving the first-visit order, yielding a fixation-ordered
sequence of unique visited patches
$\mathcal{S}=(s_1,\ldots,s_L)$. This sequence defines the causal
prediction order used by the autoregressive world model.

\paragraph{Fixation embedder.}
For each visited patch $s_i$, the fixation embedder combines the
visual patch token $\mathbf{z}_{s_i}$ with three fixation-specific
signals: a two-dimensional spatial embedding of the patch location,
a learned temporal-rank embedding for the fixation order $i$, and a
projected dwell-duration feature. The resulting fixation token is
used as input to the causal predictor.

\paragraph{Autoregressive predictor.}
The autoregressive predictor is an 8-layer causal Transformer with
12 attention heads, head dimension 64, hidden dimension $d=768$, and
a lower-triangular causal attention mask. It receives previously
fixated tokens and predicts the semantic representation of the next
fixated patch. The prediction head is a two-layer MLP with GELU
activation.

\paragraph{Momentum target encoder.}
To provide stable representation-space targets, we maintain a target
encoder $\bar{\phi}$ as an exponential-moving-average copy of the
online visual encoder $\phi$. The target encoder is evaluated under
no-gradient computation and is not directly updated by
back-propagation. After each optimizer step, its parameters are
updated as
\begin{equation}
  \bar{\theta}_{\phi}
  \leftarrow
  \tau \bar{\theta}_{\phi}
  +
  (1-\tau)\theta_{\phi},
\end{equation}
where $\theta_{\phi}$ and $\bar{\theta}_{\phi}$ denote the parameters
of the online and target visual encoders, respectively. The EMA
momentum follows a cosine schedule from $0.998$ to $1.0$.

\paragraph{Spatial-completion decoder.}
The spatial-completion branch is a lightweight 2-layer cross-attention
decoder with hidden dimension $d=768$. Let
$\mathcal{U}=\{1,\ldots,N\}\setminus\{s_1,\ldots,s_L\}$ denote the
set of unvisited patches. For each $p \in \mathcal{U}$, the decoder
forms a query by adding a learnable mask token to the spatial
embedding of patch $p$. The keys and values are the gaze-context
tokens produced by the causal predictor. The decoder outputs one
predicted representation $\hat{\mathbf{r}}_p$ for each unvisited
patch, which is matched to the corresponding momentum-target
representation $\bar{\mathbf{z}}_p$.

\paragraph{Pretraining objective.}
The world-model pretraining objective combines autoregressive
next-fixation prediction and spatial completion:
\begin{equation}
  \mathcal{L}
  =
  \mathcal{L}_{\mathrm{AR}}
  +
  \lambda
  \mathcal{L}_{\mathrm{SC}}.
\end{equation}
In all main experiments, we set the spatial-completion weight to
$\lambda = 1.0$.
The autoregressive loss is
\begin{equation}
  \mathcal{L}_{\mathrm{AR}}
  =
  \frac{1}{L-1}
  \sum_{i=2}^{L}
  \mathrm{SmoothL1}
  \left(
    \mathrm{LN}(\hat{\mathbf{z}}_{s_i}),
    \mathrm{sg}(\bar{\mathbf{z}}_{s_i})
  \right),
\end{equation}
and the spatial-completion loss is
\begin{equation}
  \mathcal{L}_{\mathrm{SC}}
  =
  \frac{1}{|\mathcal{U}|}
  \sum_{p \in \mathcal{U}}
  \mathrm{SmoothL1}
  \left(
    \mathrm{LN}(\hat{\mathbf{r}}_{p}),
    \mathrm{sg}
    \left(
      \mathrm{LN}(\bar{\mathbf{z}}_{p})
    \right)
  \right).
\end{equation}
Here, $\mathrm{LN}(\cdot)$ denotes layer normalization and
$\mathrm{sg}(\cdot)$ denotes stop-gradient.

\paragraph{Optimization.}
We train the world model on MIMIC-EYE using
AdamW~\citep{loshchilov2017decoupled} with learning rate
$3 \times 10^{-4}$, weight decay $0.04$, batch size 32, automatic
mixed precision, and 15 training epochs. The online MedCLIP-Swin
encoder, fixation embedder, causal predictor, autoregressive
prediction head, and spatial-completion decoder are optimized jointly.
The target encoder receives no gradients and is updated only through
EMA.

\paragraph{Scanpath decoder.}
GazeSearch train split: $3{,}870$ samples.
AdamW, learning rate $5 \times 10^{-4}$, weight decay 0.01, cosine
annealing ($\eta_\text{min} = 10^{-6}$, $T_\text{max} = 30$),
batch size 16, 30 epochs.
Best checkpoint selected by validation mean Euclidean distance;
evaluation on 488 test samples.

\paragraph{Linear probing.}
We evaluate frozen pretrained representations on CheXpert (14-class),
RSNA~Pneumonia, and SIIM-ACR~Pneumothorax at $1\%$, $10\%$, and
$100\%$ label fractions. For each image we extract a $1536$-d
feature by concatenating two halves: (i) the mean-pooled patch
tokens of the online MedCLIP-Swin encoder $\phi$, and (ii) the
$d$-dimensional output of a learnable readout token
prepended to a deterministic raster-scan surrogate fixation sequence
over the $N$ patches and processed by the (frozen) causal predictor,
followed by a learned linear projection back to $d$. No gaze
annotations are used at probing time. Features are standardised per
dimension (zero mean, unit variance) and fed to an
$\ell_2$-regularised logistic-regression classifier (\textsc{lbfgs},
$C{=}1$, $2000$ iterations); training fractions are sampled with a
fixed seed. We report $14$-class macro AUROC on CheXpert and binary
AUROC on RSNA / SIIM-ACR. The backbone and predictor receive no
gradient at probing time.

\begin{table}[h]
  \centering
  \caption{
    Baseline comparison protocol.
  ``Reported'' indicates numbers taken from the cited paper under
  its original or unified evaluation protocol; ``Rerun'' indicates our
  own execution. For our method, the input resolution follows the
  corresponding evaluation setting: $224$ for classification and
  $448$ for scanpath probing, to match the protocol used in each
  experiment.
  }
  \label{tab:baseline_protocol}
  \small
  \renewcommand{\arraystretch}{1.15}
  \begin{tabular}{lllll}
    \toprule
    \textbf{Method} & \textbf{Backbone} & \textbf{Pretrain data}
      & \textbf{Resolution} & \textbf{Source} \\
    \midrule
    BioViL       & ResNet    & MIMIC-CXR         & $224$  & Rerun  \\
    MedKLIP      & ResNet    & MIMIC-CXR         & $224$  & Rerun  \\
    MGCA         & ResNet    & MIMIC-CXR         & $224$  & Rerun  \\
    GLoRIA       & ResNet    & CheXpert          & $224$  & Rerun  \\
    PRIOR        & ResNet    & MIMIC-CXR         & $224$  & Rerun  \\
    MedCLIP      & Swin    & MIMIC-CXR         & $224$  & Rerun  \\
    CheXWorld    & ViT       & MIMIC-CXR         & $224$  & Rerun \\
    EGMA         & Swin    & MIMIC-CXR + EYE   & $224$  & Rerun  \\
    \midrule
    GazeFormer   & ViT          & ImageNet + COCO   & $224$  & Rerun  \\
    HAT          & ResNet    & ImageNet          & $224$  & Reported (GazeSearch) \\
    GazeSearch   & ResNet    & ImageNet          & $448$  & Rerun  \\
    LogitGaze-Med & ResNet / CheX & MIMIC-CXR + EYE  & $448$  & Reported (LogitGaze) \\
    \midrule
    Ours         & Swin    & MIMIC-CXR + EYE   & $224/448$  & Rerun \\
    \bottomrule
  \end{tabular}
\end{table}

% ----- Heatmap Visualization Protocol -----
\section{Attention Heatmap Protocol}
\label{sec:appendix:heatmap}

The attention heatmaps in Figure~\ref{fig:heatmap} are generated
using Grad-CAM~\citep{selvaraju2017grad}, following the same
protocol as EGMA~\citep{ma2024eye} to ensure visual comparability.
For each image--pathology pair:
(i)~the frozen backbone used for downstream evaluation processes the
image at $224 \times 224$;
(ii)~we compute the gradient of the target pathology logit with
respect to the feature maps of the final convolutional stage
(Swin-Tiny stage~4 for our method and EGMA; the corresponding final
block for ResNet-50-based methods);
(iii)~feature maps are globally average-pooled along the channel
dimension using the gradient magnitudes as weights;
(iv)~the resulting spatial map is ReLU-activated, upsampled to
input resolution via bilinear interpolation, and min--max normalized
to $[0, 1]$.
All methods use the same Grad-CAM implementation and the same set of
test images. Grad-CAM visualizations are qualitative: they indicate
which spatial regions most influence a particular logit but do not
provide localization guarantees.

% ----- t-SNE Visualization -----

% ----- Scanpath Decoder Architecture -----
\section{Scanpath Decoder Architecture}
\label{sec:appendix:scanpath}

The input is resized to $448 \times 448$, yielding a
$14 \times 14$ patch grid ($N = 196$) for finer spatial resolution.
Patch features are projected from $d = 768$ to $d_s = 512$ via a
linear layer.
The decoder receives three inputs: the projected patch features, a
learned task embedding
$\mathbf{t} \in \mathbb{R}^{d_s}$ indicating which pathology to
search for ($t \in \{0, \ldots, 12\}$), and the initial fixation
coordinate $\mathbf{x}_0 = (0.5, 0.5)$, projected to
$d_s$ dimensions.

At each step, the current fixation token queries the full set of
patch features via cross-attention.
A causal Transformer (6 layers, 8 heads, dimension $d_s = 512$)
processes the growing fixation sequence with temporal positional
encoding.
Three output heads operate on each hidden state $\mathbf{h}_t$:
(i) a spatial head, implemented as a 196-way softmax over the patch
grid;
(ii) a duration head that regresses fixation dwell time; and
(iii) a termination head.
Following the GazeSearch evaluation protocol, the decoder emits 7
fixations at inference time. The scanpath loss is
\begin{equation}
  \mathcal{L}_\text{scan}
  =
  \mathcal{L}_\text{spatial}(\text{CE}, 196)
  +
  0.1 \cdot \mathcal{L}_\text{dur}(\ell_1)
  +
  0.1 \cdot \mathcal{L}_\text{term}(\text{BCE}).
  \label{eq:loss_scanpath}
\end{equation}
Only the decoder parameters are trained during scanpath probing; the
backbone remains frozen in this probing stage. The scanpath loss is
never back-propagated into the backbone.

\begin{figure*}[h]
  \centering
  \includegraphics[width=\textwidth]{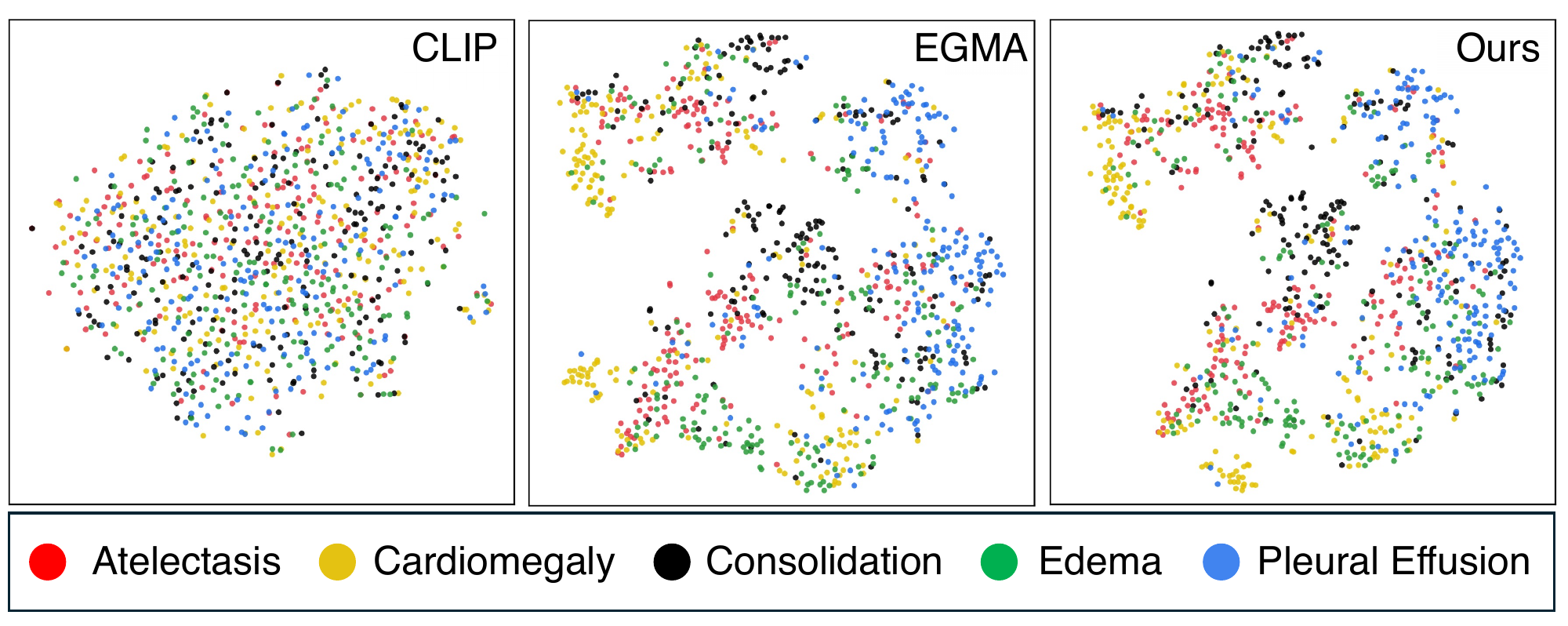}
  \caption{
    t-SNE visualization of learned representations on the CheXpert
    5$\times$200 dataset. Each point represents a chest radiograph
    and is color-coded by its ground-truth pathology label.
  }
  \label{fig:tsne}
\end{figure*}
\section{Representation Visualization}
\label{sec:appendix:tsne}

Figure~\ref{fig:tsne} provides a qualitative visualization of the
learned representation space on the CheXpert 5$\times$200 dataset.
CLIP features show considerable mixing across pathology categories,
while EGMA produces partial separation with remaining inter-class
overlap. Our method shows clearer pathology-level grouping, with more
compact clusters for categories such as Cardiomegaly and Pleural
Effusion. Since t-SNE~\citep{van2008visualizing} is a qualitative projection and can be
sensitive to hyperparameters, we use this visualization only as
supporting evidence; the quantitative transfer results in
Tables~\ref{tab:supervised} and~\ref{tab:zeroshot} provide the
primary evaluation of representation quality.